\pdfoutput=1
\documentclass[11pt]{article}
\usepackage[preprint]{acl}
\usepackage{times}
\usepackage{latexsym}
\usepackage[T1]{fontenc}
\usepackage[utf8]{inputenc}
\usepackage{microtype}
\usepackage{inconsolata}
\usepackage{hyperref}
\usepackage{graphicx}
\usepackage{booktabs}
\usepackage{textcomp}
\usepackage{multicol}
\usepackage{caption}
\title{RKadiyala at SemEval-2024 Task 8: Black-Box Word-Level Text Boundary Detection in Partially Machine Generated Texts}
\author{Ram Mohan Rao Kadiyala \\
  University of Maryland , College Park \\
  \href{mailto:rkadiyal@terpmail.umd.edu}{\texttt{rkadiyal@terpmail.umd.edu}} \\}
\begin{document}
\maketitle
\begin{abstract}
With increasing usage of generative models for text generation and widespread use of machine generated texts in various domains, being able to distinguish between human written and machine generated texts is a significant challenge. While existing models and proprietary systems focus on identifying whether given text is entirely human written or entirely machine generated, only a few systems provide insights at sentence or paragraph level at likelihood of being machine generated at a non reliable accuracy level, working well only for a set of domains and generators. This paper introduces few reliable approaches for the novel task of identifying which part of a given text is machine generated at a word level while comparing results from different approaches and methods. We present a comparison with proprietary systems , performance of our model on unseen domains' and generators' texts. The findings reveal significant improvements in detection accuracy along with comparison on other aspects of detection capabilities. Finally we discuss potential avenues for improvement and implications of our work. The proposed model is also well suited for detecting which parts of a text are machine generated in outputs of Instruct variants of many LLMs. 
\end{abstract}

\section{Introduction}
With rapid advancements and usage of AI models for text generation , being able to distinguish machine generated texts from human generated texts is gaining importance. While existing models and proprietary systems like GLTR \cite{DBLP:journals/corr/abs-1906-04043}, ZeroGPT \cite{AITextDetector}, GPTZero \cite{tian2023gptzero}, GPTKit \cite{gptkit}, Open AI detector , etc.. focus on detecting whether a given text is entirely AI written or entirely human written , there was less advancement in detecting which parts of a given text are AI written in a partially machine generated text. While some of the above mentioned systems provide insights into which parts of the given text are likely AI generated , these are often found to be unreliable and having an accuracy close or worse than random guessing. There is also a rise in usage of AI to spread fake news and misinformation along with using AI models to modify Wikipedia articles \cite{vice2023aiwikipedia}. Our proposed model focuses on detecting word level text boundary in partially machine generated texts as part of the SemEval shared task : Multi-generator, Multi-domain, and Multilingual Black-Box Machine-Generated Text Detection\cite{wang-EtAl:2024:SemEval20245}. This paper also discusses implications of findings , comparisons with different models and approaches , comparison with existing proprietary systems with relevant metrics , other findings regarding AI generated texts. The official submission is DeBERTa-CRF , several other models have been tested for comparison. With new, better, and diverse AI models coming into existence, having a model that can make accurate predictions on unseen domains and unseen generator texts can be useful for practical scenarios. 

\section{Dataset}
\begin{table}[ht]
\centering
\begin{tabular}{|l|c|c|c|}
\hline
\textbf{Set} & \textbf{Count} & \textbf{Sources} & \textbf{Generators} \\
\hline
Train & 3649 & PeerRead & ChatGPT \\
\hline
Dev & 505 & PeerRead & ChatGPT \\
\hline
Test & 11123 & PeerRead & LLaMA2 \\
\cline{3-4}
&  & OUTFOX & LLaMA2 \\
\cline{3-4}
&  & OUTFOX & GPT-4 \\
\hline
\end{tabular}
\caption{Dataset sources and split}
\label{table:1}
\end{table}
The dataset used is part of M4GT-bench  Dataset\cite{wang2024m4gtbench} consisting of texts each of which are partially human written and partially machine generated sourced from PeerRead reviews and outfox student essays \cite{koike2023outfox} all of which are in English. The generators used were GPT-4\cite{openai2024gpt4} , ChatGPT , LLaMA2 7/13/70B \cite{touvron2023llama}. \autoref{table:1} shows the source , generator used and data split of the dataset. The generators were given partially human written essays or partially human written reviews along with problem statements and instructions to complete the text. The proportion of human written content in each of the samples ranged from 0 to 50\% in the first part while the rest is machine generated in the training data and varying from 0 to 100\% in development and test sets. The length of the texts varied between a single sentence to over 20 with median word count of 212 and mean word count of 248.

\section{Baseline}
The provided baseline uses finetuned Longformer over 10 epochs. The baseline classifies tokens individually as human or machine generated and then maps the tokens to words to identify the text boundary between machine generated and human written texts. The final predictions are the labels of words after whom the text boundary exists. The detection criteria is first change from 0 to 1 or vice versa. We have tried one more approach by considering the change only if consecutive tokens are the same. The baseline model achieved an MAE of 3.53 on the Development set which consists of same source and generator as the training data. The model had an MAE of 21.535 on the test set which consists of unseen domains and generators. 

\section{Proposed Model}
\begin{figure}[ht]
\centering
\includegraphics[width=0.45\textwidth]{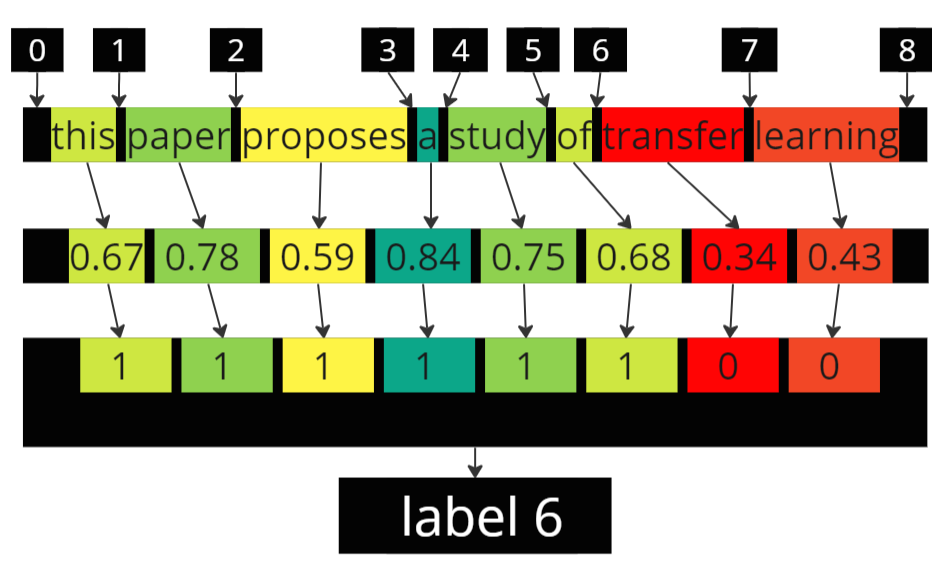}
\caption{A visual example of working of the model}
\label{figure:1}
\end{figure}

We have built several models out of which DeBERTa-CRF was used as the official submission. We have finetuned DeBERTa\cite{he2023debertav3}, SpanBERT\cite{joshi2020spanbert}, Longformer\cite{beltagy2020longformer}, Longformer-pos (Longfomer trained only on position embeddings), each of them again along with Conditional Random Fields (CRF)\cite{mccallum2012efficiently} with different text boundary identification logic by training on just the training dataset and after hyperparameter tuning , the predictions have been made on both development and test sets. CRFs have played a vital role in improving the performance of the models due to their architecture being well suited for pattern recognition in sequential data. The primary metric used was Mean Average Error (MAE) between predicted word index of the text boundaries and the actual text boundary word index. However Mean Average Relative Error (MARE) too was used for a better understanding which is the ratio of MAE and text lenght in words. Some of the plots and information couldn't be added due to page limits and are available \verb|here|. \footnote{more information available at : \url{https://www.rkadiyala.com/papers}} along with the  \verb|code used|. \footnote{Code available at : \url{https://github.com/1024-m/NAACL-2024-SemEval-TASK-8C}}. a hypothetical example in \autoref{figure:1} demonstrates how the model works. The tokens are classified at first and mapped to words. In cases where part of a word is predicted as human and rest as machine (in case of longer words), the word as a whole is classified as machine generated. 

\subsection{Our system}
We have used 'deberta-v3-base' along with CRF using Adam\cite{kingma2017adam} optimizer over 30 epochs with a learning rate of 2e-5 and a weight decay of 1e-2 to prevent overfitting. other models that have been used are 'Spanbert-base-cased', 'Longformer-base-4096', 'Longformer-base-4096-extra.pos.embd.only' which is similar to Longformer but pretrained to preserve and freeze weights from RoBERTa\cite{liu2019roberta} and train on only the position embeddings. The large variants of these have also been tested however the base variants have achieved better performance on both the development and testing datasets. predictions have been made on both the development and testing datasets by training on just the training dataset. Two approaches were used when detecting text boundary 1) looking for changes in token predictions i.e from 1 to 0 or 0 to 1. and 2) looking for change to consecutive tokens i.e 1 to 0,0 or 0 to 1,1. Approach 2 achieved better results than approach 1 in all the cases and was used in the official submission. 

\subsection{Results}
The results from using different models with the two approaches on the development set and the test set can be seen in \autoref{table:2}. These models have been trained over 30 epochs and the best results were added among the several attempts with varying hyperparameters. The provided baseline however has been trained on just through approach I over 10 epochs using base variant of Longformer.
\begin{table*}
\centering
\begin{tabular}{|l|cc|cc|cc|cc|}
\hline
\textbf{Dataset \textrightarrow} & \multicolumn{4}{c|}{\textbf{Dev set} (Seen Generator)} & \multicolumn{4}{c|}{\textbf{Test set} (Unseen Generator)} \\
\hline
\textbf{Model \textdownarrow} & \multicolumn{2}{c|}{\textbf{MAE}} & \multicolumn{2}{c|}{\textbf{MARE}} & \multicolumn{2}{c|}{\textbf{MAE}} & \multicolumn{2}{c|}{\textbf{MARE}} \\
\hline
\textbf{approach \textrightarrow} & \textbf{I} & \textbf{II} & \textbf{I} & \textbf{II} & \textbf{I} & \textbf{II} & \textbf{I} & \textbf{II} \\
\hline
DeBERTa & 3.217 & 3.174 & 0.0190 & 0.0185 & 22.031 & 19.347 & 0.1013 & 0.1006 \\
\textbf{DeBERTa-CRF} & 2.311 & \textbf{2.192} & 0.0127 & \textbf{0.0124} & 20.074 & \textbf{18.538} & 0.0919 & 0.0906 \\
SpanBERT & 6.593 & 5.918 & 0.0234 & 0.0221 & 28.406 & 25.229 & 0.1283 & 0.1274 \\
SpanBERT-CRF & 4.855 & 4.519 & 0.0196 & 0.0191 & 24.283 & 20.97 & 0.1216 & 0.1209 \\
Longformer & 3.52 & 2.878 & 0.0168 & 0.0162 & 25.985 & 21.177 & 0.1285 & 0.1103 \\
Longformer-CRF & 2.782 & 2.41 & 0.0142 & 0.0139 & 20.941 & 18.943 & 0.0964 & 0.0959 \\
Longformer.pos & 3.296 & 3.075 & 0.0177 & 0.0174 & 23.219 & 19.502 & 0.1029 & 0.1022 \\
Longformer.pos-CRF & 2.613 & 2.406 & 0.0137 & 0.0135 & 20.223 & 18.542 & 0.0911 & \textbf{0.0902} \\
\hline
Longformer (baseline) & 3.53 & & & & 21.535 & & & \\
\hline
\end{tabular}
\caption{Performance of different models and approaches on dev and test sets}
\label{table:2}
\end{table*}
These models have then been used to make predictions on the test set without further training or changes using the set of hyperparameters that produced the best results for each on the development set. However MAE which is the primary metric of the task doesn't take length of the text into consideration, Hence MARE (Mean Average Relative Error) was also calculated for a better understanding. 

\section{Comparison with proprietary systems}
Some of the proprietary systems built for the purpose of detecting machine generated text provide insights into what parts of the text input is likely machine generated at a sentence / paragraph level. Many of the popular systems like GPTZero, GPTkit, etc.. are found to to less reliable for the task of detecting text boundary in partially machine generated texts. Of the existing models only ZeroGPT was found to produce a reliable level of accuracy. For the purpose of accurate comparison percentage accuracy of classifying each sentence as human / machine generated is used as ZeroGPT does detection at a sentence level.

\subsection{Results comparison}
Since the comparison is being done at a sentence level, In cases where actual boundary lies inside the sentence, calculation of metrics is done on the remaining sentences, and when actual boundary is at the start of a sentence , all sentences were taken into consideration. With regard to predictions, A sentence prediction is deemed correct only when a sentence that is entirely human written is predicted as completely human written and vice versa. The two metrics used were average sentence accuracy which is average of percentage of sentences correctly calculated in each input text, and overall sentence accuracy which is percentage of sentences in the entire dataset accurately classified. The results on the development and test sets are as shown in \autoref{table:3}. Since its difficult to do the same on 12000 items of the test set , a small section of 500 random samples were used for comparison and were found to perform similar to the development set with a 15-20 percent lower accuracy than the proposed models. Since ZeroGPT's API doesn't cover sentence level predictions , they have been manually calculated over the development set and can be found \verb|here|. \footnote{ZeroGPT annotations available at : \url{https://docs.google.com/spreadsheets/d/1DOgAZBWQ3G6JtslQwgg9tJiX1WyZt0ajMrr2I9-yfHU/edit?usp=sharing}}.

\begin{table}[!ht]
\begin{tabular}{|l|c|c|}
\hline
\textbf{Dev set} & & \\
\hline
\textbf{Model} & \textbf{Accuracy} & \textbf{Avg. Acc..} \\
\hline
\textbf{DeBERTa-CRF} & \textbf{0.9883} & \textbf{0.9848} \\
Longformer.pos-CRF & 0.9806 & 0.9778 \\
ZeroGPT & 0.8086 & 0.7976 \\
\hline
\textbf{Test set} & & \\
\hline
\textbf{Model} & \textbf{Accuracy} & \textbf{Avg. Acc..} \\
\hline
\textbf{DeBERTa-CRF} & \textbf{0.9969} & \textbf{0.9974} \\
Longformer.pos-CRF & 0.9889 & 0.9901 \\
\hline
\end{tabular}
\caption{Performance at sentence level across Development and Test Sets}
\label{table:3}
\end{table}

\section{Conclusion}
The metrics from \autoref{table:3} demonstrate the proposed model's performance on both seen domain and generator data (dev set) along with unseen domain and unseen generator data (test set) , hinting at wider applicability. While there was a drop in accuracy at a word level, there was an increase in sentence level accuracy. 

\subsection{Strengths and Weaknesses}
It was observed that the proprietary systems used for comparison struggled with shorter texts. i.e when the input text has fewer sentences, the predictions were either that the input text is fully human written or fully machine generated leading to comparatively low accuracy.

\begin{figure}[!ht]
\centering
\includegraphics[width=0.45\textwidth]{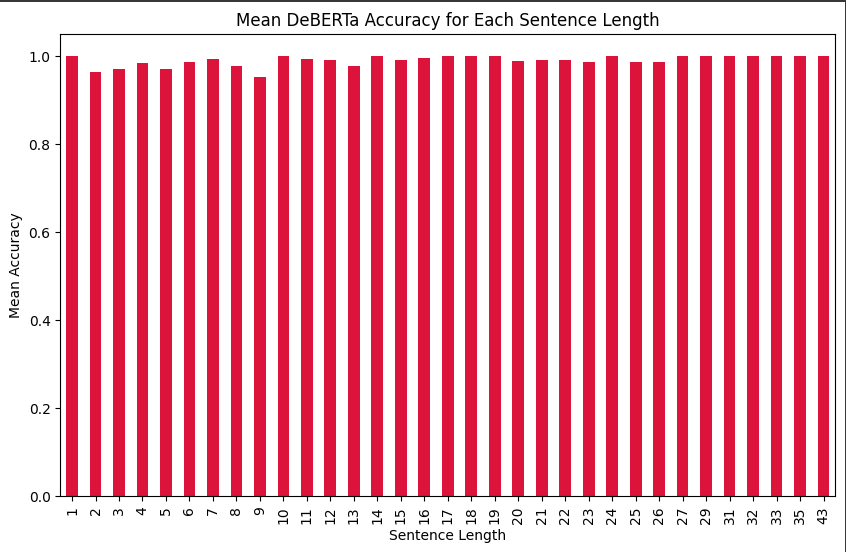}
\caption{Average sentence accuracy VS number of sentences in test set : DeBERTa-CRF}
\label{figure:2}
\end{figure}

\begin{figure}[!ht]
\centering
\includegraphics[width=0.45\textwidth]{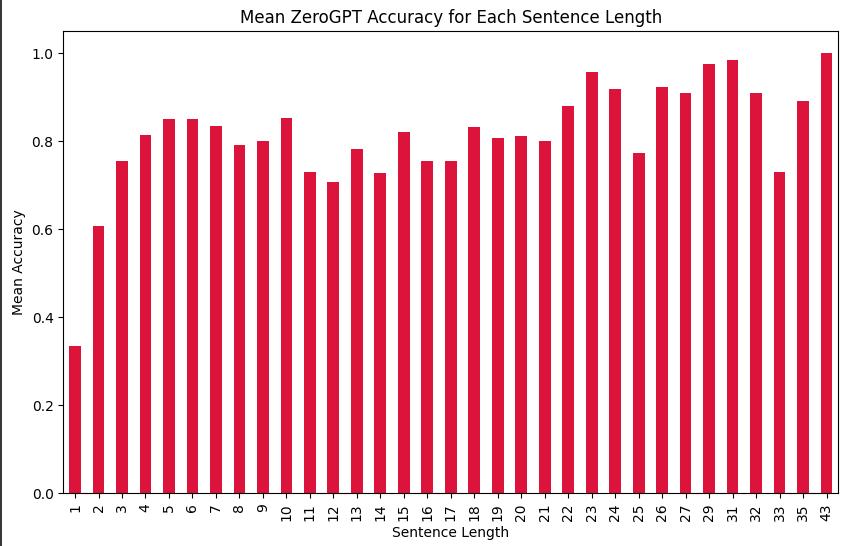}
\caption{Average sentence accuracy VS number of sentences in test set : ZeroGPT}
\label{figure:3}
\end{figure}

The average accuracy of sentence level classification for each text length of our model and ZeroGPT can be seen in \autoref{figure:2} , \autoref{figure:3} respectively. the proposed model overcomes this issue by providing more accurate results even on short text inputs. 

The sentence level accuracy did vary considerably while comparing cases where the actual text boundary is at the end of sentence and those where it is mid sentence. The results can be seen in \autoref{table:4}.

\begin{table}[ht]
\begin{tabular}{|l|c|c|}
\hline
\textbf{Model \textdownarrow} & \textbf{mid sent..} & \textbf{end of sent..} \\
\hline
DeBERTa-CRF & 0.9835 & \textbf{0.9972} \\
Longformer.pos-CRF & 0.9765 & \textbf{0.9901} \\
ZeroGPT & 0.7942 & \textbf{0.8296} \\
\hline
\end{tabular}
\caption{Performance of models based on text boundary placement : test set (approach 2)}
\label{table:4}
\end{table}
Since the source and generators of texts individually wasn't made available, the comparison between in-domain and out-of-domain texts couldn't be made.

\subsection{Possible Improvements}
\begin{figure}[!ht]
\centering
\includegraphics[width=0.45\textwidth]{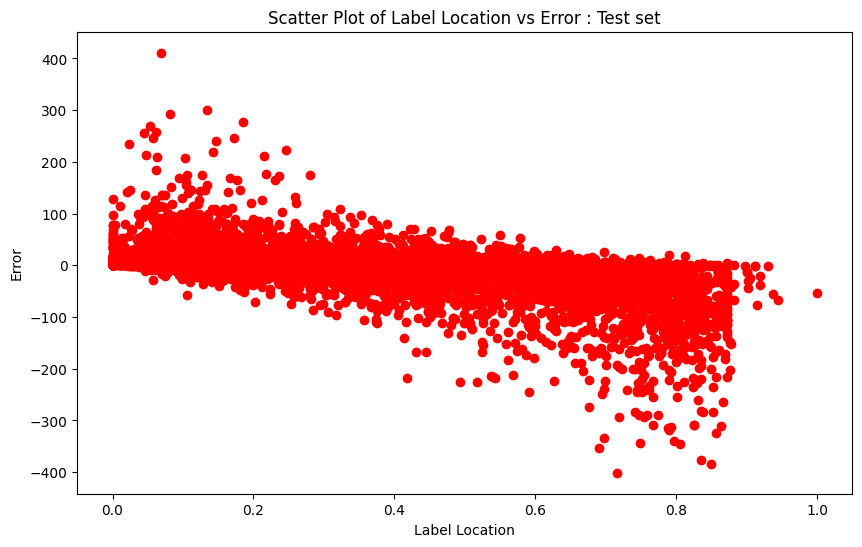}
\caption{Text boundary location VS MAE in test set : DeBERTa}
\label{figure:4}
\end{figure}

\begin{figure}[!ht]
\centering
\includegraphics[width=0.45\textwidth]{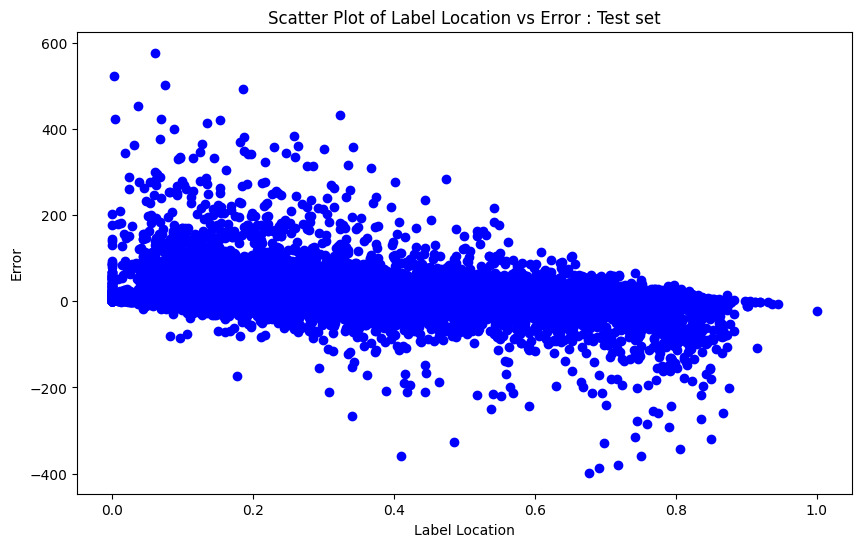}
\caption{Text boundary location VS MAE in test set : Longformer}
\label{figure:5}
\end{figure}

\begin{figure*}[!ht]
\centering
\includegraphics[width=0.922\textwidth]{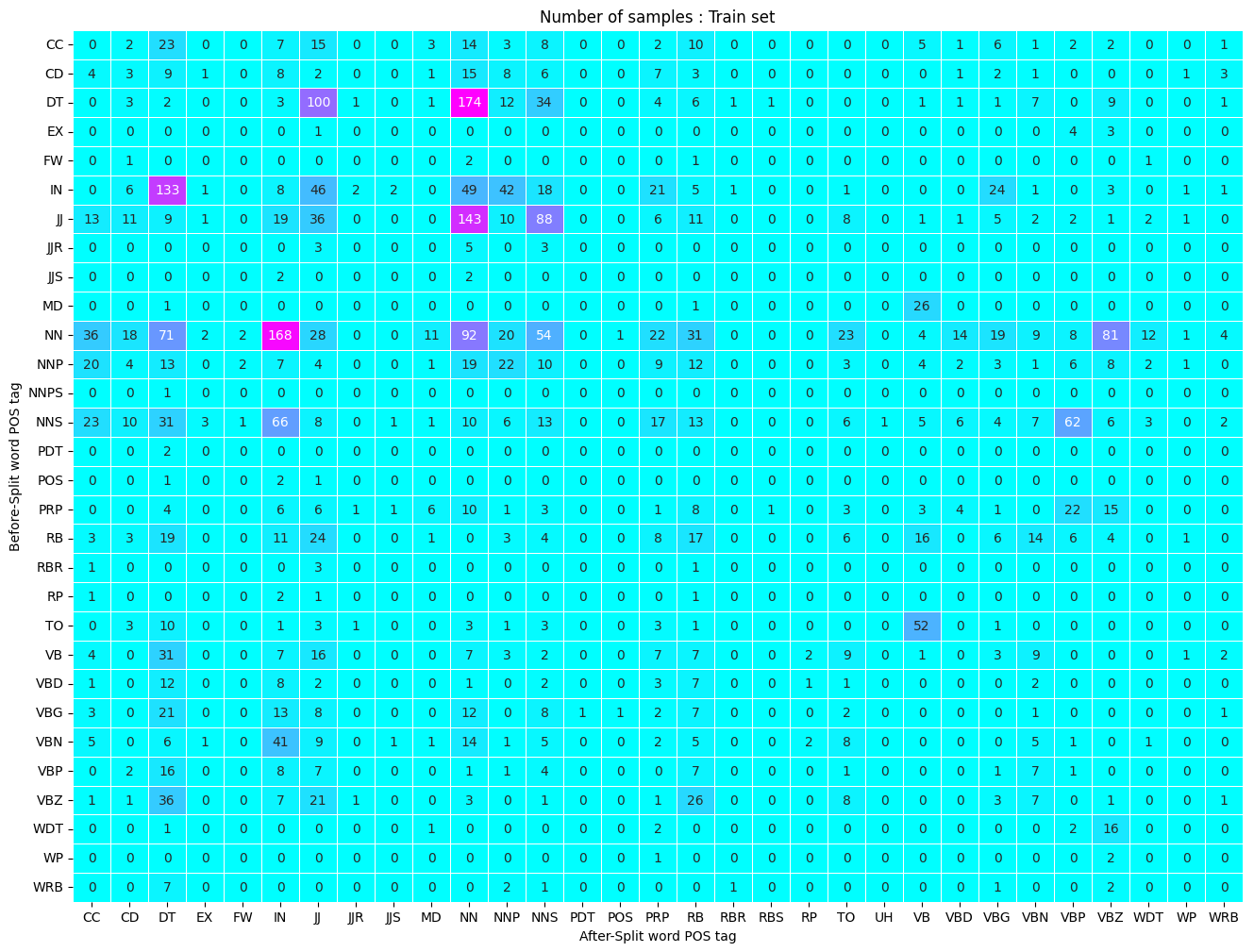}
\caption{Train set count for each pre and post text boundary POS tag combination}
\label{figure:6}
\end{figure*}

\begin{figure*}[!ht]
\centering
\includegraphics[width=0.922\textwidth]{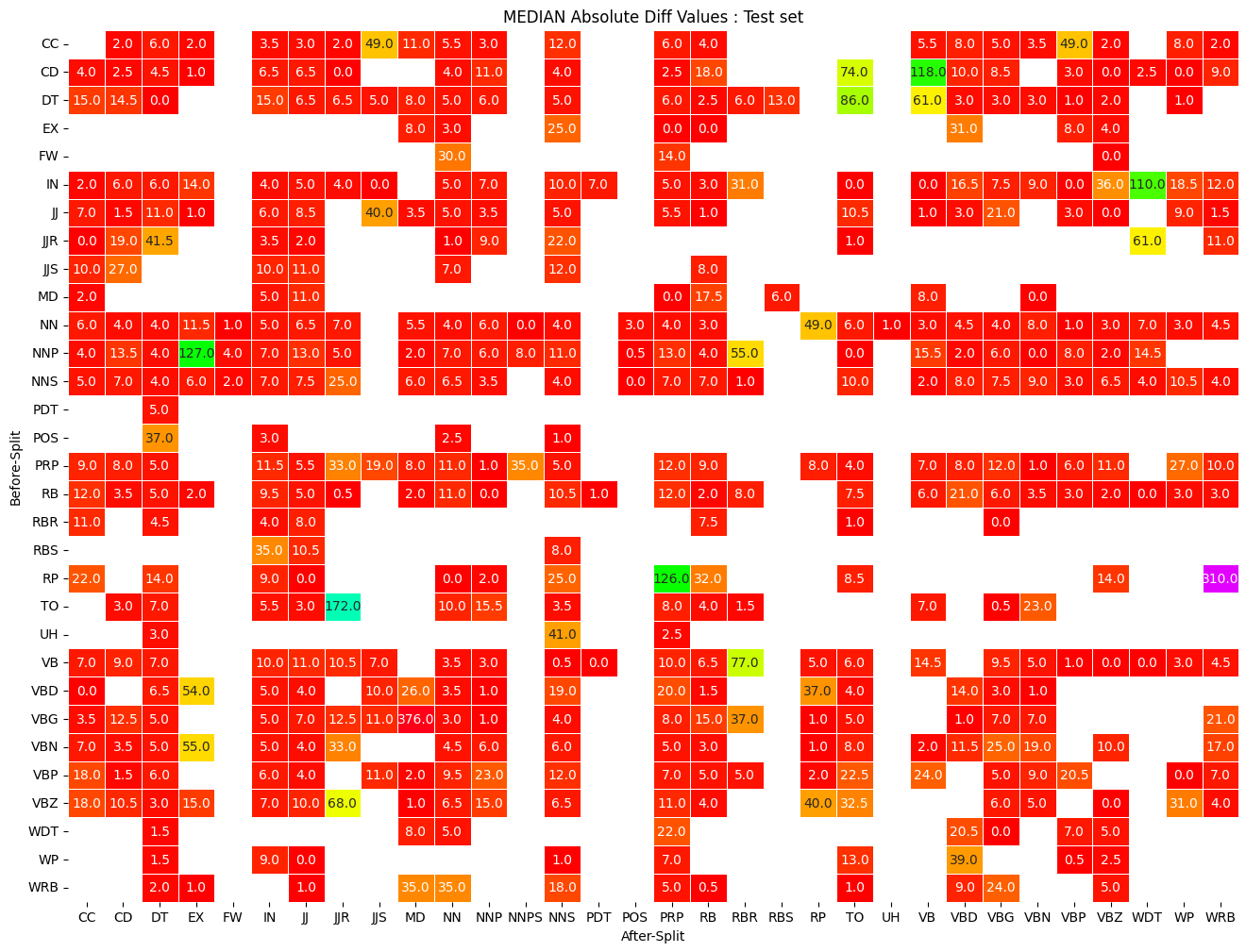}
\caption{Test set median MAE for each pre and post text boundary POS tag combination}
\label{figure:7}
\end{figure*}

DeBERTa performed better when text boundaries are in the first half of the given text, while Longformer had better performance when the text boundary is in the other half as seen in \autoref{figure:4} and \autoref{figure:5}. In cases where there was a significantly bigger MAE , atleast one of two (DeBERTa and Longformer) had made a very close prediction. There is a possibility that an ensemble of both might perform better, as seen in \autoref{table:2}, on the test set (unseen generators), while DeBERTa had a better MAE , Longformer had the better MARE. Further, the POS tags of the words pre and post text boundary were examined to find out what led to some cases having higher MAE. Though DeBERTa had better performance, when dealing with very long texts, Longformer might be a better choice. \autoref{figure:6} and \autoref{figure:7} display the count of data samples in train set and median MAE of those in test set for each POS tags combination pre and post split.
The cases where the median MAE was higher (i.e 30 or above) had none or very few samples in the training set. Excluding those cases the new MAE was less than half of what it previously was. Adding more data that covers all cases of pre-split and post-split POS tag words might lead to better results. At a sentence level the accuracy was close to 100 percent excluding the above mentioned samples. Another possible approach worth testing is having a multiplier to the predicted values of each token before classifying as a 0 or 1. 




as seen in \autoref{figure:6} and \autoref{figure:7}, the biggest error cases in pre and post text boundary POS tags were the ones which were not present at all or in very minute amount in the training data, nearly 92 percentage of cases had less than 10 samples to train on and 64 percentage of cases had no samples at all in the training set. A potential solution would be including ample amount of data for all possibilities to cover wider range of texts. This can be done through generating more data by splitting the text at required word boundaries in existing texts and using an LLM to finish the texts.

\subsection{Possible Extensions and Applications}
The need to detect AI generated content is also prevalent over all languages. While the current model utilizes just English language data, gathering multilingual data and having a multilingual model might also be of great use. With the growth of misinformation and fake news using bots on social media handles\cite{NEURIPS2019_3e9f0fc9}, being able to detect AI generated texts is of great importance. As most of the texts i.e posts , comments etc.. are shorter in length and difficult to detect, An extension of the current work by training on social media data may yield a good result as demonstrated in \autoref{figure:2} and \autoref{figure:3}. The dataset mostly consisted of texts which are academic related while there is a need to detect machine generated texts in other fields too. Also, It is worth testing the performance of paraphrased data along with the existing data. Since, usage of additional data was prohibited, data augmentation wasn't used in training the current models. It is likely that having more data to cover the cases of pre and post POS tags that weren't present in the training dataset may improve the performance of the models. Some of the other findings are available in \autoref{sec:appendix}.

\section{Limitations and potential for misuse}
While this novel task of detecting text boundaries in partially machine generated texts achieves a high accuracy where one change from human to machine occurs. Being able to handle the cases of multiple changes from human to machine and vice versa is vital. Since having a completely machine generated text and rewriting a few sentences in between or vice versa isn't covered by this work or other existing models, there is a possibility that detection can be evaded this way. There is also a potential for misuse by learning what features and texts caused errors using the proposed models to create texts that can evade detection. The current study covers only two kinds of LLMs i.e GPT and LLaMa. The performance on other types of LLMs is still to be tested. With wide range of available LLMs, training the models over wider range of LLMs might improve performance. The current work focuses on just English texts, however it can be extended to other languages by replacing DeBERTa with mDeBERTa and training on a multilingual corpus. However not all languages are covered by mDeBERTa, this can be a potential issue when dealing with multilingual texts. Another kind of texts that need to be tested upon is where machine generated portions are generated by different generators, and the cases where it is completely machine generated but by different generators. The current corpus used to trained the models is sourced from academic platforms and academic essays, It is necessary to have models to work over a wide variety of texts including cases where it can be in a casual tone. While the current work only considers the first 512 tokens, the longformer version did achieve the same results on unseen generator texts. It is worth looking into how well chunking would work on the deberta model to process longer texts.
\bibliography{custom}

\begin{thebibliography}{17}
\expandafter\ifx\csname natexlab\endcsname\relax\def\natexlab#1{#1}\fi

\bibitem[{Beltagy et~al.(2020)Beltagy, Peters, and Cohan}]{beltagy2020longformer}
Iz~Beltagy, Matthew~E. Peters, and Arman Cohan. 2020.
\newblock \href {http://arxiv.org/abs/2004.05150} {Longformer: The long-document transformer}.

\bibitem[{Gehrmann et~al.(2019)Gehrmann, Strobelt, and Rush}]{DBLP:journals/corr/abs-1906-04043}
Sebastian Gehrmann, Hendrik Strobelt, and Alexander~M. Rush. 2019.
\newblock \href {http://arxiv.org/abs/1906.04043} {{GLTR:} statistical detection and visualization of generated text}.
\newblock \emph{CoRR}, abs/1906.04043.

\bibitem[{GptKit()}]{gptkit}
GptKit.
\newblock {GPTKit: A Toolkit for Detecting AI Generated Text}.
\newblock \url{https://gptkit.ai/}.
\newblock Accessed: 2024-02-12.

\bibitem[{He et~al.(2023)He, Gao, and Chen}]{he2023debertav3}
Pengcheng He, Jianfeng Gao, and Weizhu Chen. 2023.
\newblock \href {http://arxiv.org/abs/2111.09543} {Debertav3: Improving deberta using electra-style pre-training with gradient-disentangled embedding sharing}.

\bibitem[{Joshi et~al.(2020)Joshi, Chen, Liu, Weld, Zettlemoyer, and Levy}]{joshi2020spanbert}
Mandar Joshi, Danqi Chen, Yinhan Liu, Daniel~S Weld, Luke Zettlemoyer, and Omer Levy. 2020.
\newblock \href {https://arxiv.org/abs/1907.10529} {Spanbert: Improving pre-training by representing and predicting spans}.
\newblock In \emph{Transactions of the Association for Computational Linguistics}, volume~8, pages 64--77. MIT Press.

\bibitem[{Kingma and Ba(2017)}]{kingma2017adam}
Diederik~P. Kingma and Jimmy Ba. 2017.
\newblock \href {http://arxiv.org/abs/1412.6980} {Adam: A method for stochastic optimization}.

\bibitem[{Koike et~al.(2023)Koike, Kaneko, and Okazaki}]{koike2023outfox}
Ryuto Koike, Masahiro Kaneko, and Naoaki Okazaki. 2023.
\newblock \href {http://arxiv.org/abs/2307.11729} {Outfox: Llm-generated essay detection through in-context learning with adversarially generated examples}.

\bibitem[{Liu et~al.(2019)Liu, Ott, Goyal, Du, Joshi, Chen, Levy, Lewis, Zettlemoyer, and Stoyanov}]{liu2019roberta}
Yinhan Liu, Myle Ott, Naman Goyal, Jingfei Du, Mandar Joshi, Danqi Chen, Omer Levy, Mike Lewis, Luke Zettlemoyer, and Veselin Stoyanov. 2019.
\newblock \href {http://arxiv.org/abs/1907.11692} {Roberta: A robustly optimized bert pretraining approach}.

\bibitem[{McCallum(2012)}]{mccallum2012efficiently}
Andrew McCallum. 2012.
\newblock \href {http://arxiv.org/abs/1212.2504} {Efficiently inducing features of conditional random fields}.

\bibitem[{OpenAI(2024)}]{openai2024gpt4}
OpenAI. 2024.
\newblock \href {http://arxiv.org/abs/2303.08774} {Gpt-4 technical report}.

\bibitem[{Tian and Cui(2023)}]{tian2023gptzero}
Edward Tian and Alexander Cui. 2023.
\newblock \href {https://gptzero.me} {Gptzero: Towards detection of ai-generated text using zero-shot and supervised methods}.

\bibitem[{Touvron et~al.(2023)Touvron, Lavril, Izacard, Martinet, Lachaux, Lacroix, Rozière, Goyal, Hambro, Azhar, Rodriguez, Joulin, Grave, and Lample}]{touvron2023llama}
Hugo Touvron, Thibaut Lavril, Gautier Izacard, Xavier Martinet, Marie-Anne Lachaux, Timothée Lacroix, Baptiste Rozière, Naman Goyal, Eric Hambro, Faisal Azhar, Aurelien Rodriguez, Armand Joulin, Edouard Grave, and Guillaume Lample. 2023.
\newblock \href {http://arxiv.org/abs/2302.13971} {Llama: Open and efficient foundation language models}.

\bibitem[{Vice(2023)}]{vice2023aiwikipedia}
Vice. 2023.
\newblock {AI Is Tearing Wikipedia Apart}.
\newblock \url{https://www.vice.com/en/article/v7bdba/ai-is-tearing-wikipedia-apart}.
\newblock Accessed: 2024-02-12.

\bibitem[{Wang et~al.(2024{\natexlab{a}})Wang, Mansurov, Ivanov, Su, Shelmanov, Tsvigun, Afzal, Mahmoud, Puccetti, Arnold, Aji, Habash, Gurevych, and Nakov}]{wang2024m4gtbench}
Yuxia Wang, Jonibek Mansurov, Petar Ivanov, Jinyan Su, Artem Shelmanov, Akim Tsvigun, Osama~Mohanned Afzal, Tarek Mahmoud, Giovanni Puccetti, Thomas Arnold, Alham~Fikri Aji, Nizar Habash, Iryna Gurevych, and Preslav Nakov. 2024{\natexlab{a}}.
\newblock \href {http://arxiv.org/abs/2402.11175} {M4gt-bench: Evaluation benchmark for black-box machine-generated text detection}.

\bibitem[{Wang et~al.(2024{\natexlab{b}})Wang, Mansurov, Ivanov, su, Shelmanov, Tsvigun, Mohammed~Afzal, Mahmoud, Puccetti, Arnold, Whitehouse, Aji, Habash, Gurevych, and Nakov}]{wang-EtAl:2024:SemEval20245}
Yuxia Wang, Jonibek Mansurov, Petar Ivanov, jinyan su, Artem Shelmanov, Akim Tsvigun, Osama Mohammed~Afzal, Tarek Mahmoud, Giovanni Puccetti, Thomas Arnold, Chenxi Whitehouse, Alham~Fikri Aji, Nizar Habash, Iryna Gurevych, and Preslav Nakov. 2024{\natexlab{b}}.
\newblock \href {https://aclanthology.org/2024.semeval2024-1.275} {Semeval-2024 task 8: Multidomain, multimodel and multilingual machine-generated text detection}.
\newblock In \emph{Proceedings of the 18th International Workshop on Semantic Evaluation (SemEval-2024)}, pages 2041--2063, Mexico City, Mexico. Association for Computational Linguistics.

\bibitem[{Zellers et~al.(2019)Zellers, Holtzman, Rashkin, Bisk, Farhadi, Roesner, and Choi}]{NEURIPS2019_3e9f0fc9}
Rowan Zellers, Ari Holtzman, Hannah Rashkin, Yonatan Bisk, Ali Farhadi, Franziska Roesner, and Yejin Choi. 2019.
\newblock \href {https://proceedings.neurips.cc/paper_files/paper/2019/file/3e9f0fc9b2f89e043bc6233994dfcf76-Paper.pdf} {Defending against neural fake news}.
\newblock In \emph{Advances in Neural Information Processing Systems}, volume~32. Curran Associates, Inc.

\bibitem[{ZeroGPT()}]{AITextDetector}
ZeroGPT.
\newblock Zerogpt : Reliable chat gpt, gpt4 \& ai content detector.
\newblock \url{https://www.zerogpt.com}.

\end{thebibliography}

\appendix
\section{Other Plots and information}
\label{sec:appendix}
Some of the information that couldn't be covered due to page limitations along with details for system replication have been added here.
\subsection{POS tag usage : human vs machines}
\label{sec:appendix_A}
\begin{figure}[ht]
\centering
\includegraphics[width=0.47\textwidth]{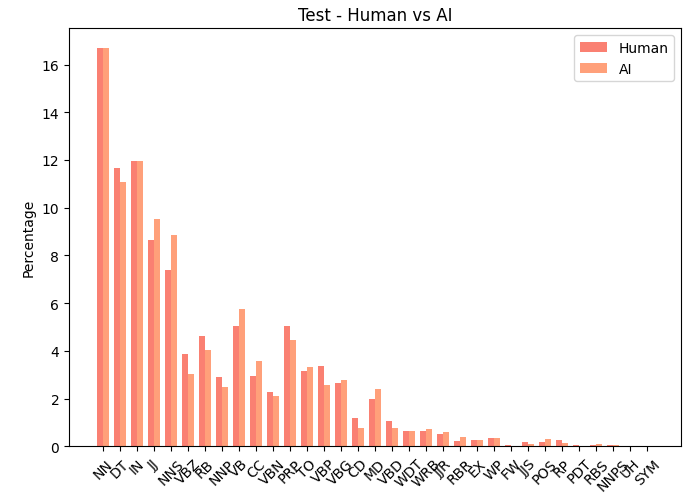}
\caption{Percentile distribution of each POS tag in test set : human VS machine}
\label{figure:10}
\end{figure}

\begin{figure}[ht]
\centering
\includegraphics[width=0.45\textwidth]{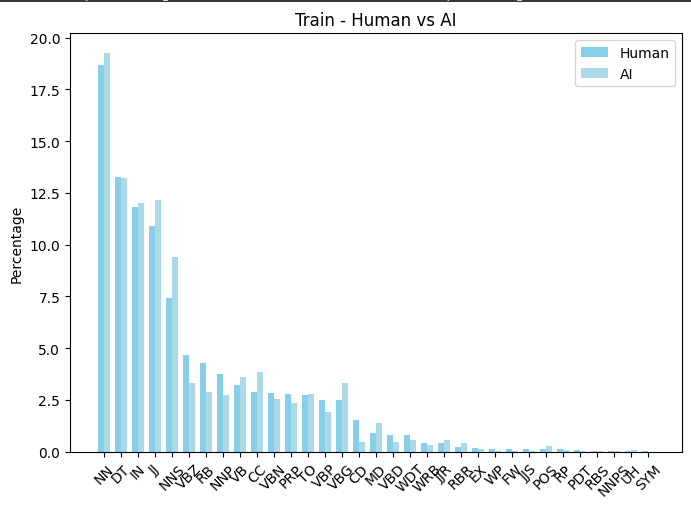}
\caption{Percentile distribution of each POS tag in train set : human VS machine}
\label{figure:8}
\end{figure}

\begin{figure}[ht]
\centering
\includegraphics[width=0.45\textwidth]{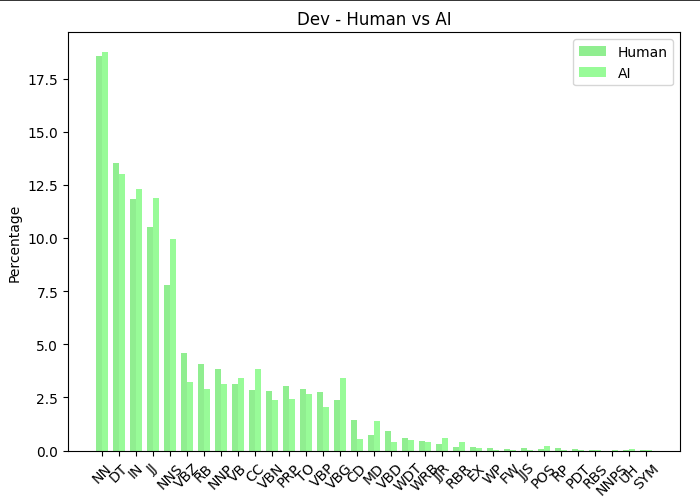}
\caption{Percentile distribution of each POS tag in dev set : human VS machine}
\label{figure:9}
\end{figure}

It can be seen from \autoref{figure:8} , \autoref{figure:9} and \autoref{figure:10} that machine generated texts had higher share of certain POS tags in the machine generated parts compared to the human written parts. This was observed in all 3 sets, the train and dev had  similar distributions as a result of using same generators i.e ChatGPT and the test had a bit of a variation due to multiple different generators i.e LLaMA2 and GPT4. Although the percentile comparison did vary from train, dev and test sets , it was minimal.

\begin{figure}[ht]
\centering
\includegraphics[width=0.45\textwidth]{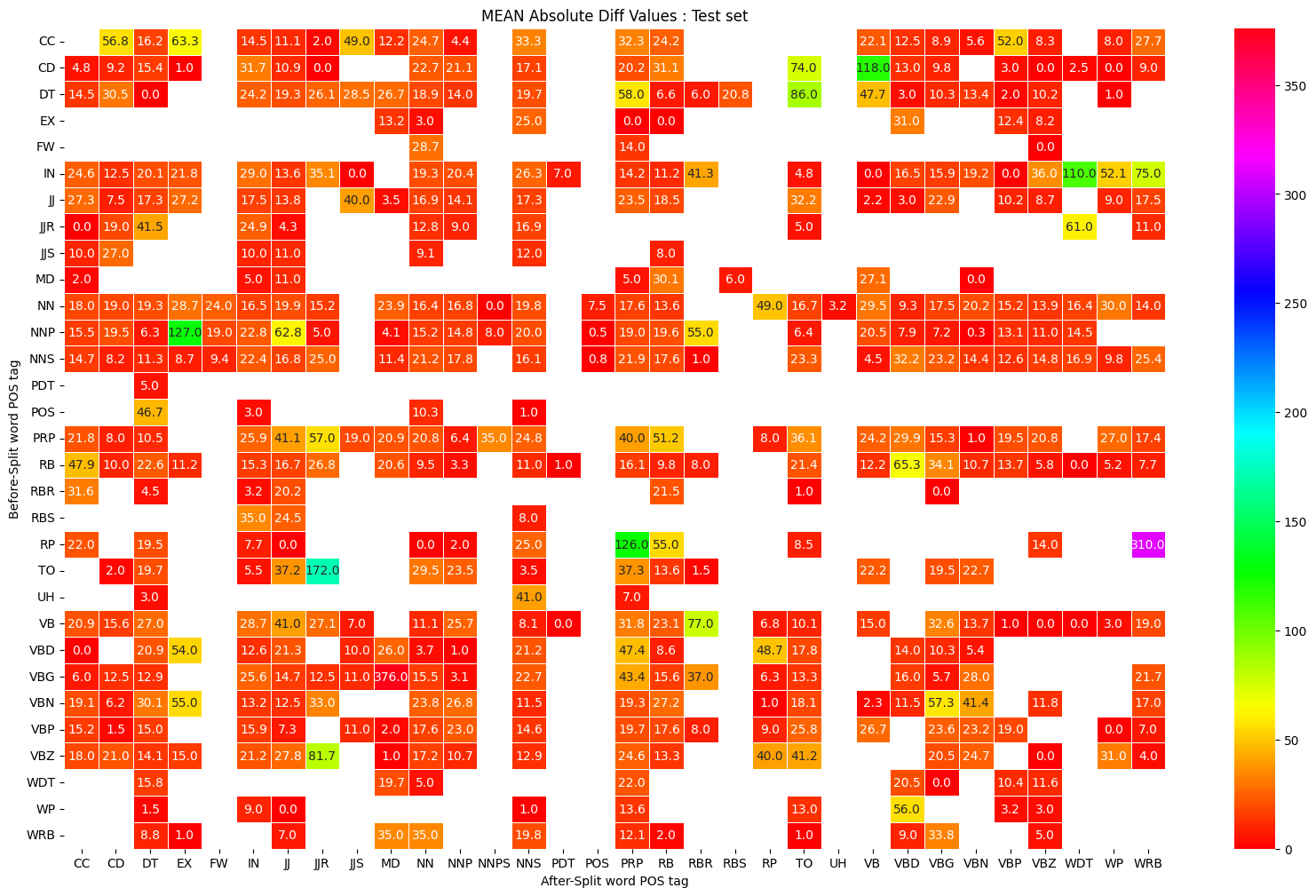}
\caption{Median MAE based on pre and post text boundary POS tags : DeBERTa-CRF}
\label{figure:11}
\end{figure}
\begin{figure}[ht]
\centering
\includegraphics[width=0.45\textwidth]{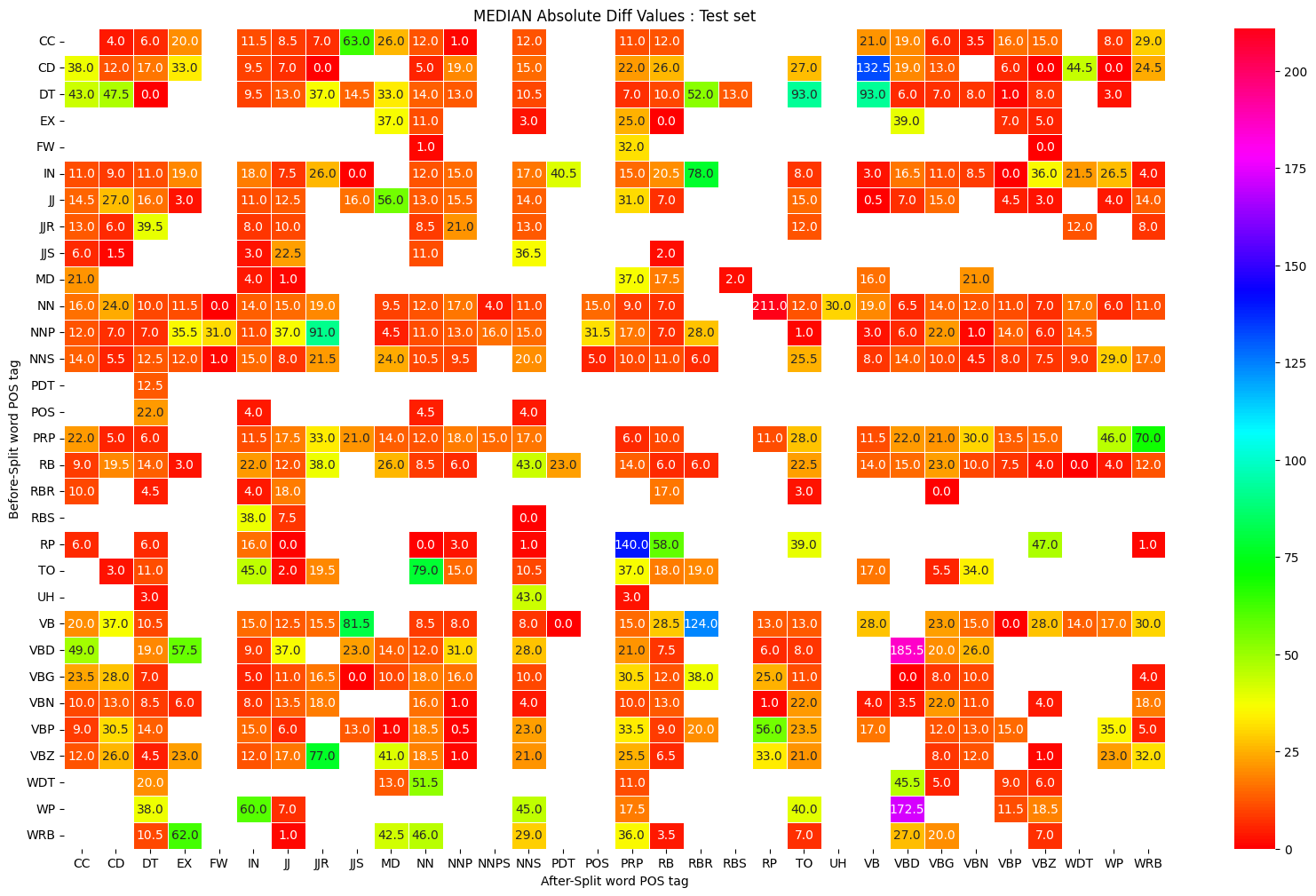}
\caption{Median MAE based on pre and post text boundary POS tags : Longformer.pos-CRF}
\label{figure:12}
\end{figure}

\subsection{MAE characteristics : DeBERTa vs Longformer}
\label{sec:appendix_A2}
As discussed in the paper , there were some instances where one model performed significantly better than the other as seen in \autoref{figure:11} and \autoref{figure:12} hinting that an ensemble of both's predictions might yield better results. 

\section{System Description}
DeBERTa-CRF was the official submission, longformer.pos-CRF had almost the same performance on the test set. i.e 18.538 and 18.542 .
\label{sec:appendix_C}
\\

\begin{table}[!ht]
\centering
\begin{tabular}{|l|l|}
\hline
\multicolumn{2}{|c|}{Official submission model configuration} \\
\hline
Base model & microsoft/deberta-\\
& -v3-base \\
\hline
\centering
Finetuning : & \\
\hspace{2ex}Learning rate & $2 \times 10^{-5}$ \\
\hspace{2ex}Weight decay & $1 \times 10^{-2}$ \\
\hspace{2ex}CRF Dropout rate & $75 \times 10^{-4}$ \\
\hspace{2ex}Max length & 512 tokens \\
\hspace{2ex}Epochs & 30 \\
\hspace{2ex}Optimizer & Adam \\
\hline
Preprocessing & No \\
\hline
Trained on & only train set \\
\hline
Sentence separation & nltk: '!' , '.' , '?' \\
\hline
Hardware & 1xV100 GPU 16GB \\
\hline
\end{tabular}
\caption{Official submission system description : DeBERTa-CRF}
\label{table:6}
\end{table}


\begin{table}[!ht]
\centering
\begin{tabular}{|l|l|}
\hline
\multicolumn{2}{|c|}{Secondary model configuration} \\
\hline
Base model & allenai/longformer-base \\
& -4096-extra.pos.embd... \\
\hline
Finetuning : & \\
\hspace{2ex}Learning rate & $2 \times 10^{-5}$ \\
\hspace{2ex}Weight decay & $1 \times 10^{-2}$ \\
\hspace{2ex}CRF Dropout rate & $1 \times 10^{-2}$ \\
\hspace{2ex}Max length & 4096 tokens \\
\hspace{2ex}Epochs & 30 \\
\hspace{2ex}Optimizer & Adam \\
\hline
Preprocessing & No \\
\hline
Trained on & only train set \\
\hline
Sentence separation & nltk: '!' , '.' , '?' \\
\hline
Hardware & 1xV100 GPU 16GB\\
\hline
\end{tabular}
\caption{Unofficial submission system description : Longformer.pos-CRF}
\label{table:7}
\end{table}

Other models that have been tested but were found to have a big margin of performance with above listed models

\begin{table}[!ht]
\centering
\begin{tabular}{|l|}
\hline
Other models tested \\
\hline
microsoft/deberta-v3-large \\
microsoft/deberta-v3-small \\
microsoft/deberta-v3-xsmall \\
SpanBERT/spanbert-base-cased \\
SpanBERT/spanbert-large-cased \\
allenai/longformer-base-4096 \\
allenai/longformer-large-4096 \\
allenai/longformer-large-4096-extra.pos.embd \\
\hline
\end{tabular}
\caption{Other models tested as part of the task}
\label{table:8}
\end{table}

Due to time and computational resources limitation, only a part of hyperparameter space was explored.

\begin{table}[!ht]
\centering
\begin{tabular}{|l|l|}
\hline
\multicolumn{2}{|c|}{Hyperparameter space explored} \\
\hline
\hspace{2ex}Learning rate & $1 \times 10^{-5}$ \\
& $2 \times 10^{-5}$ \\
& $3 \times 10^{-5}$ \\
\hline
\hspace{2ex}Weight decay & $1 \times 10^{-2}$ \\
& $2 \times 10^{-2}$ \\
& $25 \times 10^{-3}$ \\
& $5 \times 10^{-2}$ \\
\hline
\hspace{2ex}CRF Dropout rates & $2 \times 10^{-2}$\\
& $15 \times 10^{-3}$\\
& $1 \times 10^{-2}$ \\
& $90 \times 10^{-4}$\\
& $80 \times 10^{-4}$\\
& $75 \times 10^{-4}$\\
& $70 \times 10^{-4}$\\
& $60 \times 10^{-4}$\\
\hline 
\hspace{2ex}Max length & 512 tokens \\
& 4096 *longformer \\
\hline
\hspace{2ex}Epochs & 10 to 30 \\
\hline
\hspace{2ex}Optimizers &  Adafactor \\
& Adam \\
\hline
\hspace{2ex}Training data & full train set \\
& full train+dev set \\
& 80\% train set \\
\hline
\end{tabular}
\caption{Hyperparameters explored on the models}
\label{table:9}
\end{table}

\section{Effect of Text boundary location on performance}
\label{sec:appendix_D}
\begin{figure}[ht]
\centering
\includegraphics[width=0.45\textwidth]{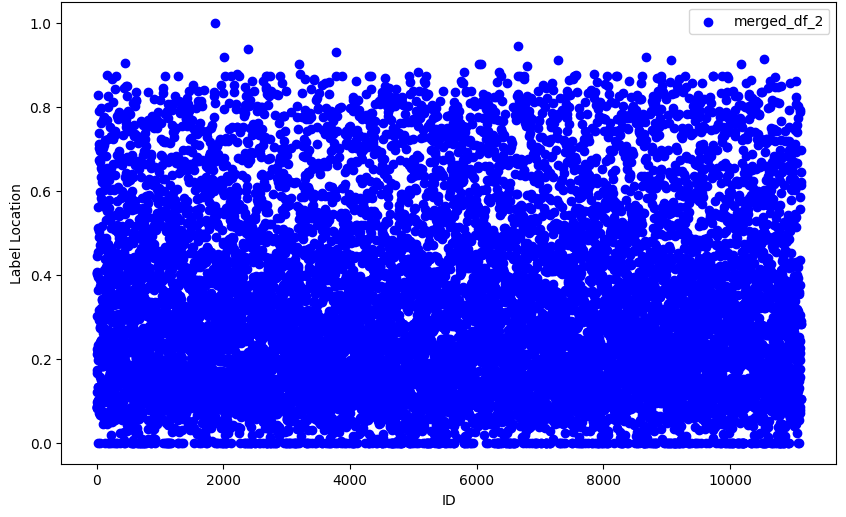}
\caption{Location of text boundary : testing set}
\label{figure:13}
\end{figure}

\begin{figure}[ht]
\centering
\includegraphics[width=0.45\textwidth]{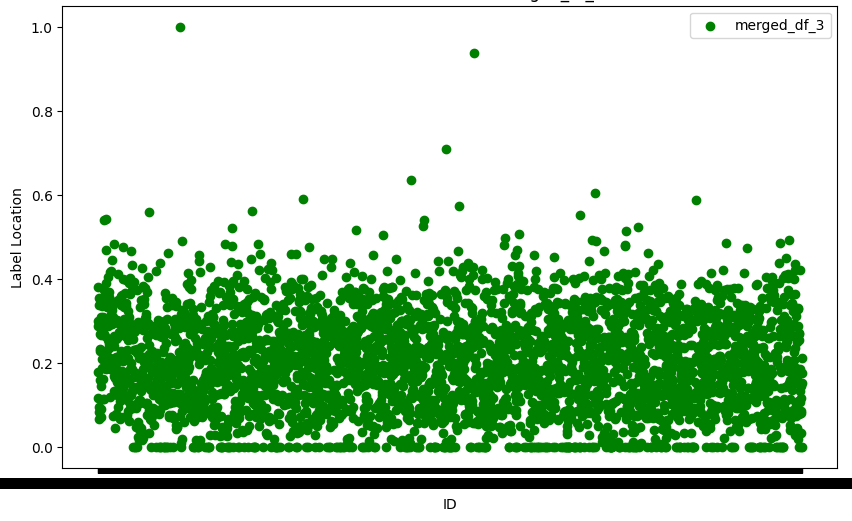}
\caption{Location of text boundary : training set}
\label{figure:14}
\end{figure}

The location of text boundaries with respect to length of the text samples are varying over the training and testing set as seen in \autoref{figure:13} and \autoref{figure:14}. Despite training on samples where the text boundaries are in the first half in most of the cases, the models did perform well on the testing set where there is a good amount of samples with text boundaries in later half. This is an area where the proprietary systems struggled. 

\end{document}